\documentclass[lettersize,journal]{IEEEtran}
\usepackage{amsmath,amsfonts}
\usepackage{array}
\usepackage[caption=false,font=normalsize,labelfont=sf,textfont=sf]{subfig}
\usepackage{textcomp}
\usepackage{stfloats}
\usepackage{url}
\usepackage{verbatim}
\usepackage{graphicx}
\usepackage{cite}
\usepackage{amssymb}
\usepackage{booktabs}
\usepackage{multirow}
\usepackage{makecell}
\usepackage{pifont}
\usepackage{enumitem}
\usepackage[ruled,linesnumbered]{algorithm2e}
\usepackage{hyperref}
\usepackage{float}

\begin{document}

\title{Fuse, Reason and Verify: Geometry Problem Solving with Parsed Clauses from Diagram}

% \author[label1,label2]{Ming-Liang Zhang}
% \ead{zhangmingliang2018@ia.ac.cn}
% \author[label2,label1]{Zhong-Zhi Li}
% \ead{lizhongzhi2022@ia.ac.cn}
% \author[label1,label2]{Fei Yin}
% \ead{fyin@nlpr.ia.ac.cn}
% \author[label3]{Liang Lin}
% \ead{linlng@mail.sysu.edu.cn}
% \author[label1,label2]{Cheng-Lin Liu\corref{cor1}}
% \ead{liucl@nlpr.ia.ac.cn}

% \affiliation[label1]{organization={MAIS, Institute of Automation of Chinese Academy of Sciences},
%             city={Beijing},
%             country={China}}
% \affiliation[label2]{organization={School of Artificial Intelligence, University of Chinese Academy of Sciences},
%             city={Beijing},
%             country={China}}
% \affiliation[label3]{organization={School of Data and Computer Science, Sun Yat-Sen University},
%             city={Guangzhou},
%             country={China}}

% \cortext[cor1]{Corresponding author.}

\author{
Ming-Liang Zhang,
Zhong-Zhi Li,
Fei Yin,
Liang Lin
and Cheng-Lin Liu
}

% \author{
% Michael~Shell,~\IEEEmembership{Member,~IEEE, }
% John~Doe,~\IEEEmembership{Fellow,~OSA,} 
% and~Jane~Doe,~\IEEEmembership{Life~Fellow,~IEEE}
% \thanks{Manuscript received January 20, 2002; revised August 26, 2015. This work was supported by the IEEE.}% 
% \thanks{M. Shell was with the Georgia Institute of Technology.}
% } 

\markboth{Journal of \LaTeX\ Class Files,~Vol.~14, No.~8, August~2021}%
{Shell \MakeLowercase{\textit{et al.}}: A Sample Article Using IEEEtran.cls for IEEE Journals}


\maketitle

\begin{abstract}
Geometry problem solving (GPS) requires capacities of multi-modal understanding, multi-hop reasoning and theorem knowledge application. In this paper, we propose a neural-symbolic model for plane geometry problem solving (PGPS), named PGPSNet-v2, with three key steps: modal fusion, reasoning process and knowledge verification. In modal fusion, we leverage textual clauses to express fine-grained structural and semantic content of geometry diagram, and fuse diagram with textual problem efficiently through structural-semantic pre-training. For reasoning, we design an explicable solution program to describe the geometric reasoning process, and employ a self-limited decoder to generate solution program autoregressively. To reduce solution errors, a multi-level theorem verifier is proposed to eliminate solutions that do not match geometric principles, alleviating the hallucination of the neural model. We also construct a large-scale geometry problem dataset called PGPS9K, containing fine-grained annotations of textual clauses, solution program and involved knowledge tuples. Extensive experiments on datasets Geometry3K and PGPS9K show that our PGPSNet solver outperforms existing symbolic and neural solvers in GPS performance, while maintaining good explainability and reliability, and the solver components (fusion, reasoning, verification) are all justified effective.
\end{abstract}

\begin{IEEEkeywords}
Geometry problem solving, Multi-modal fusion, Geometric logic reasoning, Theorem knowledge verification.
\end{IEEEkeywords}

\section{Introduction}

\IEEEPARstart{A}{utomatic} geometry problem solving (GPS) has been long-standing in artificial intelligence (AI) field~\cite{NEVINS19751,Chou2000,Seo2015,9404866}, and is drawring increasing attention in recent years~\cite{Lu2021,Chen2021,Ning2023,trinh2024solving}. A geometry problem usually consists of a geometry diagram and a textual problem, forming a multi-modal problem. The textual problem describes the geometric conditions and sets the reasoning target in natural language, while the geometry diagram displays the spatial structure and additional geometric conditions in vision. Solving geometry problems requires geometry diagram and textual problem understanding and multi-step reasoning incorporating geometry knowledge. Due to the diverse and irregular contents of textual problem and geoetry diagram, GPS is widely recognized as a vital testbed~\cite{narboux2018computer} for evaluating the high-level multi-modal reasoning capability of AI.

The challenges of GPS lie in its three main steps: modal fusion, reasoning and knowledge verification. Modal fusion is to represent and utilize multi-modal content from geometry diagram and textual problem. Fusing two distinct modalities is difficult, where the textual problem expresses geometric information by text while the geometry diagram conveys geometric relationships via visual layout. Reasoning is sophisticated involving state transition and path search~\cite{KAPUR198861,9960856}.
% as shown in Figure~\ref{fig:key-steps}. 
State transition, embedded in path search, applies geometric pattern matching and geometry theorem to move forward the state of solution. Imitating from human problem-solving, knowledge verification plays a key role in checking the solving process and correcting solution errors. 

% \begin{figure}[t]
%     \centering
%     \includegraphics[width=0.50 \textwidth]{./picture/key-steps.pdf} 
%     \caption{Reasoning process of GPS. (a) state transition; (b) path search.}
%     \label{fig:key-steps}
% \end{figure}

\begin{figure*}[t]
    \centering
    \includegraphics[width=0.90\textwidth]{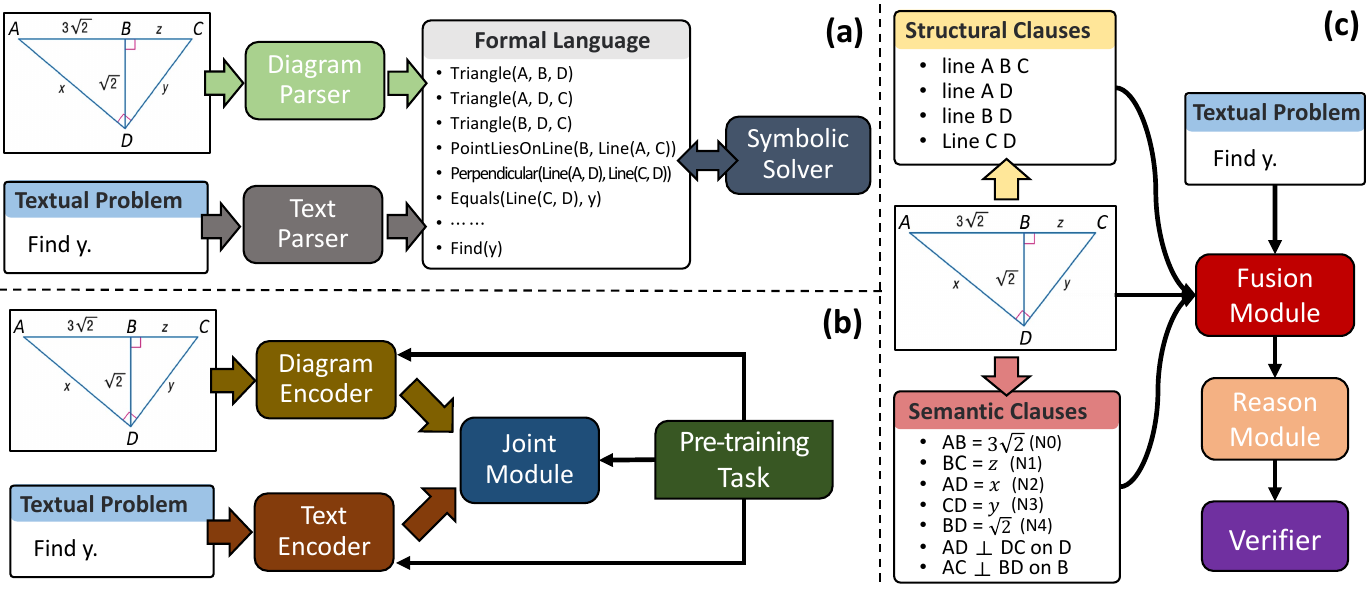} 
    \caption{Framework comparison of existing geometric solvers. (a) symbolic solvers; (b) neural solvers; (c) our PGPSNet-v2 solver.}
    \label{fig:model_contrast}
\end{figure*}

To address the above three challenges, several geometric solvers have been proposed, which can be roughly categorized into two groups: symbolic solvers and neural solvers. As summarized in Figure~\ref{fig:model_contrast} (a), symbolic solvers~\cite{Wong2009,Seo2015,Lu2021} parse the diagram and textual problem into a unified formal language, and then perform symbolic reasoning through path search and condition matching according to geometry knowledge. During the process, new conditional states are generated progressively until the search target is achieved. Although symbolic solvers have strong interpretability, they are designed with many human efforts, and may exhibit slow problem-solving speed due to redundant steps. In contrast, neural solvers~\cite{Cao2022,Chen2022,Ning2023} use neural encoders to extract features of diagram and textual problem, and embed them into a unified feature space by the joint module, as shown in Figure \ref{fig:model_contrast} (b). The neural solver is trained end-to-end in a data-driven way and outputs sequential solutions finally. Despite the promise of neural solvers, they still struggle in modal fusion and knowledge interpretability, resulting in a certain performance gap compared with symbolic solvers~\cite{Lu2021,Lu2023}. Regarding modal fusion, existing neural solvers, adopting similar frameworks of general vision-language tasks applied for natural images~\cite{Anderson2018,Yu2019,Kim2021}, cannot utilize the structural and semantic information in diagram explicitly and efficiently. The inherent black-box characteristic of neural networks also hinders the explainability and reliability of solution.

Considering the pros and cons of current geometric solvers, we propose a neural-symbolic solver, called PGPSNet-v2, as shown in Figure \ref{fig:model_contrast} (c). PGPSNet-v2 is the improved version of our previously proposed neural solver PGPSNet~\cite{Zhang2023-ijcai}. It consists of three modules: fusion, reason and verifier, to enhance the performance from three aspects of modal fusion, reasoning and knowledge verification, respectively. The fusion module inputs not only geometry diagram and textual problem, but also textual clauses parsed from geometry diagram. Multi-modal fusion is accomplished with the strategy of structural and semantic pre-training, so as to model global structure and context in unified neural form. The reason module applies a self-limited decoder to generate interpretable reasoning sequences in an autoregressive manner. The verifier validates the generated solution sequences in three levels: form, calculability and semantic, through knowledge tuple search and matching. By the way of verification in post-processing, PGPSNet-v2 excludes hallucinations that do not match geometric principles and thus promotes the reliability of solving process.

To facilitate GPS research, we also build and release a large-scale GPS dataset named PGPS9K, containing 9,021 plane geometry problems, each paired with a textual problem and a geometry diagram. PGPS9K is more complete compared with existing GPS datasets~\cite{Seo2015,Lu2021,Chen2021} because it offers both fine-grained annotations of textual clauses from diagram and solution programs, and corresponding theorem knowledge base and program executor. Textual clauses, including structural clauses and semantic clauses, possess highly syntactic and less redundant information. Given the complexity of GPS, problem-solving procedure of PGPS9K is designed as a solution program based on geometry theorems, wherein each step denotes an application of a theorem (axiom), instead of the fundamental arithmetic operations used in math word problem (MWP) datasets~\cite{Wang2017,Amini2019}. Our solution program carries richer geometry knowledge, better interpretability and shorter sequence length. The theorem knowledge base, stored in the form of knowledge tuples with operator, operands, theorem formula and semantic rules, serve the verifier to validate the solution program. Responding to the solution program, the program executor realizes symbolic algebraic calculation according to theorem formulas and returns numerical answer. In addition, we adopt five strategies of problem augmentation based on the equivalence of geometric representation, to increase the problem diversity further and embed representation knowledge simultaneously. Extensive experiments on PGPS9K and Geometry3K~\cite{Lu2021} datasets demonstrate that PGPSNet-v2 achieves remarkable improvement of GPS performance, exceeding existing symbolic solvers and neural solvers.

In summary, our main contributions are as follows: 
\begin{itemize}
    \item We propose a neural-symbolic geometric solver PGPSNet-v2 based on three core steps: fusion, reasoning and verification.
    \item We construct a large-scale plane geometry problem dataset PGPS9K with fine-grained annotations and paired with theorem knowledge tuples.
    \item The proposed PGPSNet-v2 solver generates explainable and reliable solution process, thereby yielding superior GPS performance.
\end{itemize}

The remainder of this paper is organized as follows: Section 2 reviews related works. Section 3 describes the PGPS9K dataset. Section 4 introduces the pipeline of PGPSNet-v2 solver in detail. Section 5 presents experimental results, ablation studies, case analysis and limitation discussion. Section 6 draws concluding remarks.

\section{Related Work}

\subsection{Multi-modal Reasoning}
Multi-modal reasoning task combines multi-modal information to carry out reasoning, generally in the form of questions and answers, e.g., visual question answering (VQA)~\cite{Anderson2018,Kim2021}, document question answering (DQA)~\cite{Xu2020,Tito2021} and table question answering (TQA)~\cite{lu2023dynamic,Zhu2021}. The knowledge types investigated in multi-modal reasoning include common sense~\cite{hudson2019gqa,Xu2020}, semantics~\cite{Xu2020,Lu2021a}, numerical quantity~\cite{Johnson2017,lu2023dynamic}, spatial location~\cite{lu2022learn,Johnson2017}, color~\cite{Lu2021a,hudson2019gqa}, and so on. To accomplish multi-modal reasoning, one must initially comprehend and amalgamate multi-modal content, and subsequently employ domain knowledge to address problems logically. Due to the discrepancy of data modality and reasoning mechanisms, there are significant semantic gaps among different multi-modal reasoning tasks. The GPS stands out as a unique form of multi-modal reasoning, delving into the cognition of geometric spatial structures and mathematical logical reasoning. Additionally, GPS necessitates the utilization of geometry theorem knowledge, making it more complicated. 

\subsection{Geometry Problem Solving}

Generalized GPS problems can be categorized into two types: geometric theorem proving and geometric numerical calculation. Early efforts in GPS were primarily focused on the automatic proving of geometric theorems~\cite{Nevins1975,Wong2009,trinh2024solving}, and the geometric numerical calculation~\cite{Seo2015,Sachan2017} started to gain momentum in recent years as new datasets and methods are presented. In this paper, \textit{we concentrate on the geometric numerical calculation} and review the related works on two categories: symbolic solvers and neural solvers.

Symbolic solvers~\cite{Seo2015,Lu2021,Peng2023,trinh2024solving} typically employ symbolic-based computation and rule-based deduction to solve geometry problems, featuring explicit and clear steps in the solution and reasoning process. GEOS~\cite{Seo2014,Seo2015} constructed geometric structures through a graphic parser, utilized heuristic search to expand unknown conditions, and matched option conditions to obtain the solution results. Inter-GPS~\cite{Lu2021} leveraged the diagram parser and text parser to unify modalities into formal language, and used the symbolic solver equipped with three search strategies to search and generate solutions iteratively. GeoDRL~\cite{Peng2023} enhanced the search strategy of Inter-GPS by taking GPS as a Markov decision process, and employed logical graph deduction and deep reinforcement learning methods to optimize the geometric reasoning. Nevertheless, the symbolic solvers are designed carefully with complex rules, and hard to extend to other diverse domains of geometry problems.

Neural solvers~\cite{Chen2021,Chen2022,Ning2023}, treating GPS as a special vision-language task, rely on the neural network for end-to-end learning and reasoning, and design a solution sequence to represent the solution process. NGS~\cite{Chen2021} and Geoformer~\cite{Chen2022} used auxiliary self-supervised tasks such as location prediction, elements prediction and knowledge classification to boost cross-modal semantic representation. DPE-NGS~\cite{Cao2022} boosted the NGS model and embedded a bidirectional parallel text encoder (DPE) to make long-text question encoding more efficient. SCA-GPS~\cite{Ning2023} tried to align character in text and diagram and enhanced the diagram understanding through multi-label classification and masked image modeling pre-training. However, current neural solvers are still coarse-grained in modal understanding and fusion especially for geometry diagrams, and not trustworthy for solutions with massive hallucinations. 

Our proposed PGPSNet-v2 is an improved version of neural sovler, combining modal fusion, reasoning process and knowledge verification to overcome the shortages of existing methods. It combines the strengths of neural sovler and symbolic solver, thus can be viewed as a neural-symbolic solver. It also utilizes geometry knowledge in both implicit and explicit ways to improve the interpretability and reliability of solution.

%\subsection{Knowledge Representation}

%Knowledge representation~\cite{tian2022knowledge,ji2021survey} is a crucial step for reasoning, and various schemes of explicit and implicit knowledge representation have been proposed. In the era of symbolic learning, the formal language~\cite{Bobrow1964} and knowledge graph~\cite{hogan2021knowledge} are widely adopted to express knowledge. These formalized representation can precisely articulate concepts, properties, relationships, as well as the processes of deduction and proof~\cite{10236457}. However, explicit knowledge usually relies on extensive manual compilation, and knowledge formulation may become intricate and inflexible when comes to high-level logical knowledge~\cite{ji2021survey}. In contrast, the current language models, usually based on the Transformer architecture~\cite{Vaswani2017} implicitly acquire knowledge through extensive pre-training~\cite{Devlin2019,Lewis2020} on large corpora. This representation scheme is highly flexible, and exhibits superior performance in various downstream reasoning tasks~\cite{Khashabi2020,han2021pre} when provided with a sufficient amount of pre-training data. However, this kind of implicit knowledge is influenced and limited by data distribution, and has been criticized for its opacity and uncontrollability~\cite{Patel2021}. Our PGPSNet-v2 solver represents geometry knowledge in both implicit and explicit ways, implicitly fusing structure and context with structural-semantic pre-training, and explicitly verifying candidate solutions with the theorem knowledge tuples.

\section{PGPS9K Dataset}

% where visual primitives, e.g. point, line, circle, text and symbol, are related with each other structurally and semantically

\subsection{Collection and Statistics}

\begin{table*}[t]
    \caption{Comparison with existing GPS datasets. Type, KB, OP and PL refer to problem type, knowledge base, operator number and program length, respectively.}
    \centering
    % \scriptsize
    % \renewcommand\arraystretch{1.2}
    \begin{tabular}{lccccccccc}
        \toprule
        Dataset    & \#QA & Grade & \#Type & Diagram Anno & Rationale & KB  & \#Avg OP & \#Avg PL  \\
        \midrule
        GEOS \cite{Seo2015} & 186  & 6-10 & - & No & - & No & - & - \\
        GEOS++ \cite{Sachan2017} & 1,406 & 6-10 & - & No & - & No & - & - \\
        GEOS-OS \cite{Sachan2017a} & 2,235 & 6-10 & - & No & Demonstration & No & - & - \\
        Geometry3K \cite{Lu2021} & 3,002 & 6-12 & 4 & Yes & Logical form  & No & - & - \\
        GeoQA \cite{Chen2021} & 4,998 & 6-12 & 3 & No & Program & No & 1.98 & 5.35 \\
        GeoQA+ \cite{Cao2022} & 7,528 & 6-12 & 3 & No & Program & No & 2.61 & - \\
        UniGeo \cite{Chen2022} & 9,543 & 6-12 & 5 & No & Proving Sequence & No & - & - \\
        PGPS9K & 9,021 & 6-12 & 30 & Yes & Program & Yes & 2.43 & 7.45  \\
        \bottomrule
    \end{tabular}

    \label{tab:dataset_compare}
\end{table*}

Most existing datasets of GPS, as recorded in Table~\ref{tab:dataset_compare}, either have a small data size, suitable only for rule-based symbolic solvers, or own coarse-grained annotations, neglecting the rich information in geometry diagram. To promote the development of GPS field, we construct a large-scale GPS dataset, named PGPS9K\footnote{\url{http://www.nlpr.ia.ac.cn/databases/CASIA-PGPS9K}}, annotated with fine-grained diagram labels (elements and relationships) and interpretable solving solution, along with the corresponding theorem knowledge base and program executor. To the best of our knowledge, PGPS9K is the largest and most comprehensive GPS dataset up to date. It consists of 9,021 textual problems illustrated with 4,000 non-repetitive geometry diagrams. Among them, 2,891 textual problems and 1,738 geometry diagrams were selected from the Geometry3K dataset~\cite{Lu2021}, while the remaining samples were collected from five geometry textbooks\footnote{\url{https://www.mheducation.com/}} across 6-12 grades. The samples of PGPS9K are categorized into 30 problem types based on geometry knowledge points, covering almost all types of plane geometry problems. More statistical details are exhibited in Appendix \ref{Problem Types}. 

\begin{figure*}[t]
    \centering
    \includegraphics[width=0.78\textwidth]{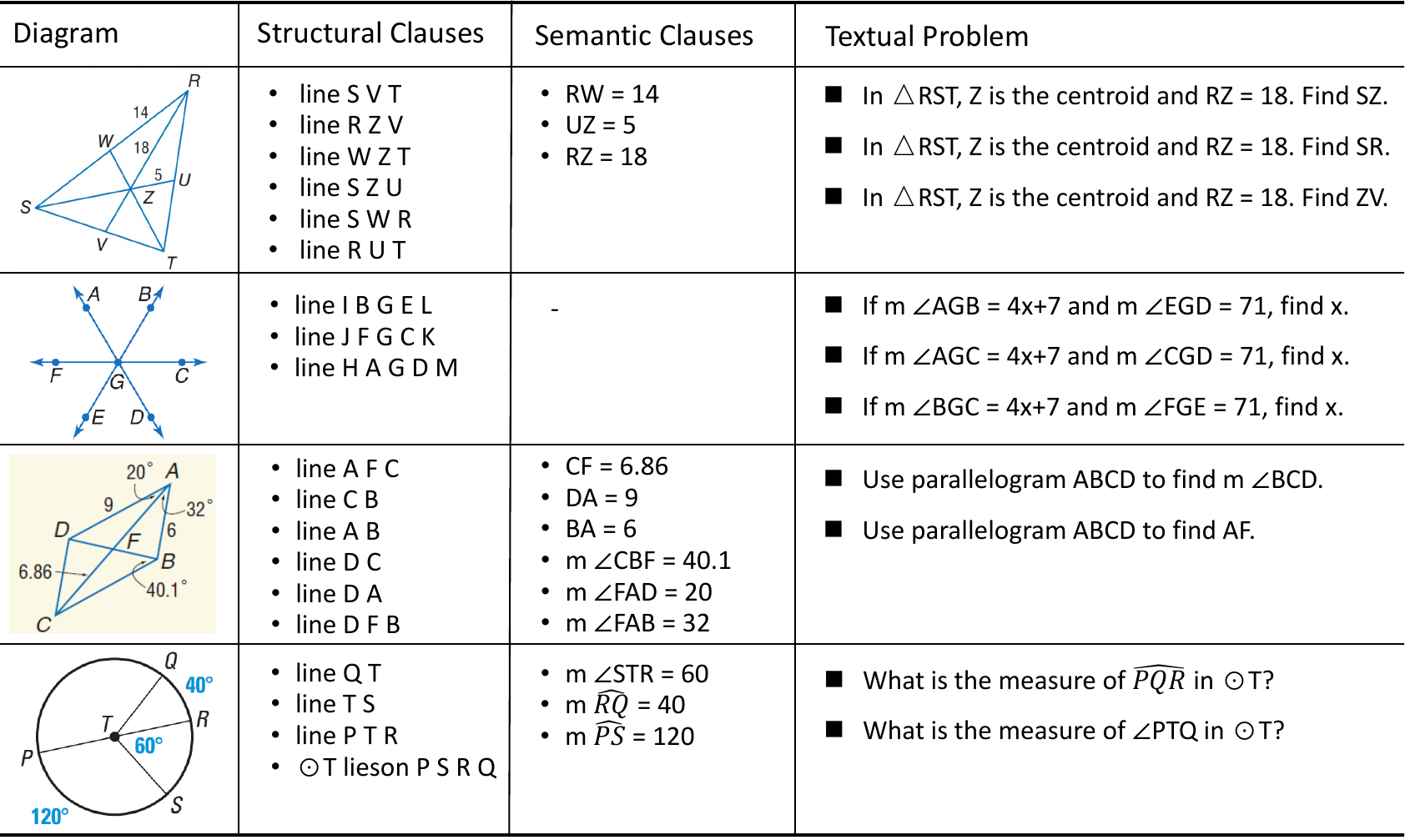} 
    \caption{Some examples of geometry problems in our PGPS9K dataset.}
    \label{fig:problem_example}
\end{figure*}

Furthermore, as shown in Figure~\ref{fig:problem_example}, PGPS9K possesses five properties, enabling it focus on the challenges in geometric reasoning and alleviate the bias introduced by the text~\cite{Manjunatha2019,Patel2021}. The properties are: (1) \textit{Theorem-based}: In the GPS process, it is necessary to use geometric theorem or axiom knowledge to carry out algebraic calculation, and finally give the numerical result; (2) \textit{Diagram-dependent}: More than 90\% of problems must be solved in conjunction with diagram, because necessary conditions such as numerical content and geometric structure are presented by visual diagram instead of textual problem; (3) \textit{Abstract}: Geometry diagram is composed of basic geometric primitives (point, line, circle) and non-geometric primitives (text, symbol). No complex semantic scenarios are involved in textual problem except abstract geometric conditions; (4) \textit{Fine-grained}: Problems with the same diagram may vary in conditions or targets. Slight distinctions in textual problems usually lead to completely different solutions; (5) \textit{Condition-redundancy}: Some conditions in textual problem or diagram are not necessarily used in GPS. The statistical results on PGPS9K dataset show that an average of 1.9 conditions are not used for GPS, 42\% of problems have redundant conditions, and 15\% of problems have 3 or more unused conditions.

\subsection{Annotation and Description}

The annotations of PGPS9K dataset include diagram annotation and solution procedure, wherein the diagram annotation serves to articulate the structural and semantic information within diagrams, and the solution procedure delineates the steps for solving geometry problems.

The diagrams employ primitive-level annotation consistent with the geometry diagram parsing \cite{Zhang2022}, containing geometric/non-geometric primitives as well as relationships among primitives in tuple form. The annotated primitives and relationships can be transformed into two fundamental types of textual clauses: structural clauses (three types) and semantic clauses (six types), via simple clause templates matching. The structural clauses depict the connectivity relationships among geometric primitives, such as points on a line or a circle, wherein points are arranged in a certain order. The connectivity relationships reveal the most fundamental geometric structural connections, often depicted in diagram but frequently omitted in textual problem. The semantic clauses articulate the basic relationships between geometric primitives and non-geometric primitives in natural language. The relationships corresponding to semantic clauses are essential components for GPS, complementing with textual problem. More details about textual clauses are given in Appendix \ref{Textual Clauses}. It is worth mentioning that the definition of textual clauses remains open and extensible. The overarching design principle is to describe the complete structural-semantic information of diagram for assisting GPS. 

\begin{figure}[t]
    \centering
    \includegraphics[width=0.45 \textwidth]{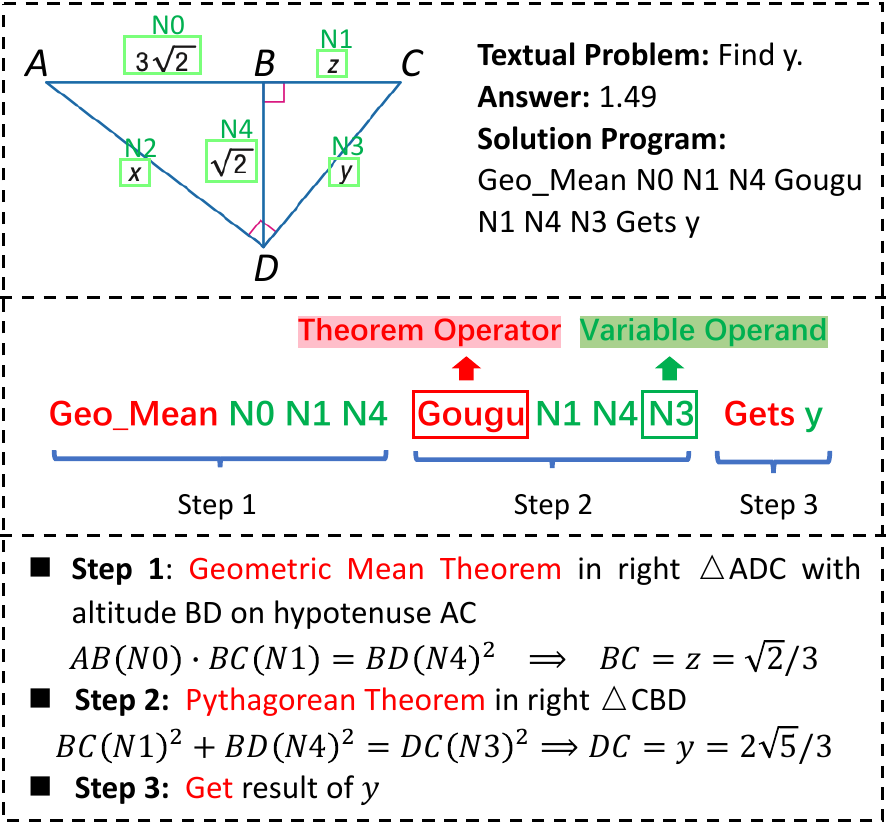} 
    \caption{Annotation design of solution program and its interpretability.}
    \label{fig:annotation}
\end{figure}

The problem-solving procedure of PGPS9K is designed as a solution program composed of multiple deductive steps. The command set of solution program consists of 34 operators $O\!P$ and 55 operands $P\!N$, wherein operands encompass 11 problem variables $N$ (appearing in textual problem or semantic clauses), 7 intermediate variables $V$ (generated during solving process), 26 augments $A\!R\!G$ (letter unknowns), and 11 constants $C$. As shown in Figure~\ref{fig:annotation}, one solving step is formed by one operator and multiple operands, involving a geometric theorem or axiom, with relevant operands arranged in the semantic order of theorem formula. For example, within a right triangle, the Geometric Mean Theorem reveals the length relation between altitude $c$ and two segments $a, b$ into which the altitude divides the hypotenuse, with the theorem formula $a*b=c^2$, so we express it as ``Geo\_Mean(a,b,c)”. In comparison to other annotation schemes~\cite{Amini2019,Tsai2021,Chen2021}, our annotation omits basic arithmetic operations such as +, -, *, /, instead replacing them with theorem operations, exhibiting advantages in terms of structure, knowledge guidance and interpretability (see Figure \ref{fig:annotation}). It makes the solving procedure more concise and mitigates the difficulty of model learning. Furthermore, for the first time, we introduce the intermediate variable $V$ as unknown variables within steps but also as transfer variables cross steps, to unify forward and backward computation operations. For instance, in the Pythagorean Theorem, ``Gougu(V,*,*)" and ``Gougu(*,*,V)" can be set to solve the length of legs and hypotenuse, respectively. More details of solution program are demonstrated in Appendix \ref{Solution Program}.

\subsection{Knowledge Base and Program Executor}

For knowledge matching of geometry problem, PGPS9K dataset is equipped with the knowledge base of plane geometry theorems expressed in the form of knowledge tuples. One knowledge tuple consists of four elements: operator, operands, theorem formula and semantic rules, as displayed in Table~\ref{tab:knowledge_tuple}. The operators are abbreviations of theorems or axioms of totally 34 types. The operands are arranged according to the semantic order of theorem formula, and several theorem operands can be expressed in multiple ways, such as ``Ratio" (scale ratio), ``Circle\_R\_Circum" (circumference of circle or arc), ``Circle\_D\_Area" (area of circle or sector), etc. The formulas of theorems are given and the relevant semantic rules are defined in tuples. Here are three specific examples of knowledge tuple:
\begin{itemize}
    \item ``(Gougu, (a,b,c), $a^2+b^2=c^2$, ...)" is the knowledge tuple corresponding to the Pythagorean Theorem, where (a,b,c) corresponds to two right legs and hypotenuses, the theorem formula is $a^2+b^2=c^2$ and its semantic rules are $a+b>c, 0<a<c, 0<b<c$.
    \item ``(Gcos, (a,b,c), $cos(c) = a/b$, ...)" is the knowledge tuple corresponding to the Cosine Law, where (a,b,c) corresponds to two sides and an angle respectively, the theorem formula is $cos(c) = a/b$, the defined semantic rules are $0<c<90, 0<a<b$, and in a right triangle, $a$ and $b$ are adjacent sides and $b$ is the hypotenuse.
    \item ``(Kite\_Area, (a,b,c), $a* b/2=c$, ...)" is the knowledge tuple of kite-shaped area, where (a,b,c) corresponds to two diagonal sides and area respectively, the theorem formula is $a* b/2=c$, and the semantic rules are $a,b,c>0$ and corresponding lines of $a$ and $b$ are intersect.
\end{itemize}

\begin{table*}[t]
\caption{Geometry theorem knowledge tuples. Noting that semantic rules are omitted due to limited space.}
\centering
\begin{tabular}{lll|lll} 
\toprule
\textbf{Operator}          & \textbf{Operands}   & \textbf{Formula}  & \textbf{Operator}          & \textbf{Operands}   & \textbf{Formula}               \\ 
\midrule
Get               & (a)              & Get value of a  & Equal             & (a, b)           & a = b                                            \\ 
% \midrule
Sum               & (a, b, c, …, d) & a + b + c + … = d & Multiple          & (a, b, c, …, d) & a * b * c * … = d                                \\ 
% \midrule
Iso\_Tri\_Ang     & (a, b)           & a + 2 * b = 180 & Gougu             & (a, b, c)        & a\^{}2 + b\^{}2 = c\^{}2                         \\ 
% \midrule
Gsin              & (a, b, c)        & sin(c) = a / b  & Gcos              & (a, b, c)        & cos(c) = a / b                                   \\ 
% \midrule
Gtan              & (a, b, c)        & tan(c) = a / b  & Cos\_Law          & (a, b, c, d)     & a\^{}2 = b\^{}2 + c\^{}2 - 2 * b * c             \\ 
% \midrule
Sin\_Law          & (a, b, c, d)     & sin(a) / b = sin(c) / d & Median            & (a, b, c)        & a + c = 2 * b~~~                                 \\ 

% \midrule
 \multirow{2}{*}{Proportion}       & (a, b, c, d)     & a / b = c / d & \multirow{2}{*}{Ratio}     & (a, b, c)        & a / b = c \\
 & (a, b, c, d, e)  & (a / b)\^{}e = c / d  &
                  & (a, b, c, d)     & (a / b)\^{}c = d                     \\ 
% \midrule
 Geo\_Mean         & (a, b, c)        & a * b = c\^{}2  &  Chord2\_Ang       & (a, b, c)        & a = (b + c) / 2                                  \\ 
% \midrule
TanSec\_Ang       & (a, b, c)        & a = (c - b) / 2  &                      
Tria\_BH\_Area    & (a, b, c)        & a * b / 2 = c                                    \\
% \midrule
Tria\_SAS\_Area   & (a, b, c, d)     & a * c * sin(b) / 2 = d  &
PRK\_Perim        & (a, b, c)        & (a + b) * 2 = c                                  \\ 
% \midrule
Para\_Area        & (a, b, c)        & a * b = c &
Rect\_Area        & (a, b, c)        & a * b = c                                        \\ 
% \midrule
Rhom\_Area        & (a, b, c)        & a * b * 2 = c &
Kite\_Area        & (a, b, c)        & a * b / 2 = c                                    \\ 
% \midrule
 \multirow{2}{*}{Circle\_R\_Circum} & (a, b)           & 2 * pi * a = b  &                                 \multirow{2}{*}{Circle\_D\_Circum} & (a, b)           & pi * a = b                                       \\ 
 & (a, b, c)        & 2 * pi * a * b / 360 = c     &              
                  & (a, b, c)        & pi * a * b / 360 = c                             \\ 
 \multirow{2}{*}{Circle\_R\_Area}   & (a, b)           & pi * a\^{}2 = b  &    
 \multirow{2}{*}{Circle\_D\_Area}   & (a, b)           & pi * (a / 2)\^{}2 = b                              \\ 
                  & (a, b, c)        & pi * a \^{}2 * b / 360 = c  &
                  & (a, b, c)        & pi * (a / 2)\^{}2 * b / 360 = c                    \\ 
% \midrule
Trap\_Area        & (a, b, c, d)     & (a + b) * c / 2 = d     &                ArcSeg\_Area      & (a, b, c)        & pi * a\^{}2 * b / 360 - a\^{}2 * sin(b) / 2 = c  \\ 
% \midrule
Ngon\_Ang         & (a, b)           & (a - 2) * 180 = b    &                     RNgon\_B\_Area    & (a, b, c)        & a * b\^{}2 / tan(180/a) / 4 = c                  \\ 
% \midrule
RNgon\_L\_Area    & (a, b, c)        & a * b\^{}2 * sin(360/a) / 2 = c          & RNgon\_H\_Area    & (a, b, c)        & a * b\^{}2 * tan(180/a) = c                      \\
\bottomrule
\end{tabular}
\label{tab:knowledge_tuple}
\end{table*}

GPS not only requires geometric reasoning but also involves algebraic operations, such as solving linear and nonlinear equations, as listed in Table \ref{tab:knowledge_tuple}. Our program executor is constructed in conjunction with solution program and capable of executing solution programs following theorem formulas. It is built upon the SymPy library of Python, enabling execution of symbolic algebraic operations for one or multiple unknowns, while recording values of all variables throughout the solving process. In case of reasoning steps that cannot proceed, the executor provides corresponding explanatory prompts. For those challenging steps, particularly those involving nonlinear symbolic operations, the executor sets a time threshold to terminate computation proactively, avoiding affecting the operation of whole system. 

To sum up, the knowledge base stored via knowledge theorem tuples can express theorem knowledge explicitly, and the developed program executor exhibits enhanced functionality, adept at handling various algebraic operations related to geometry problem. As a result, they collectively establish a robust groundwork for the verifier of PGPSNet-v2 solver.

\subsection{Problem Augmentation}

% \begin{figure*}[t]
%     \centering
%     \includegraphics[width=0.7 \textwidth]{./picture/augmentation.pdf} 
%     \caption{Data augmentation of geometry problem.}
%     \label{fig:augmentation}
% \end{figure*}

Despite PGPS9K being the largest and highest-quality GPS dataset to date, it still fails to adequately fulfill the learning requirements of neural solvers due to the large variety of geometry problems. Therefore,
% as depicted in Figure \ref{fig:augmentation},
we adopt five problem augmentation strategies based on the diversity and equivalence of geometric representations to further expand the richness of PGPS9K:

\begin{itemize}
    \item \textit{Token Replacement}: The replaceable tokens include three types: points, angle IDs and arguments. Once a token is altered, all identical tokens in both textual clauses and textual problem should be uniformly replaced. For example, if point B is replaced with point V, the new text would be: ``line V C D", ``$\odot$A lieson E V D", ``VD $\perp$ EA on C" and ``Find VD".
    
    \item \textit{Connection Rotation}: By changing the order of points, the connection relationships within the structural clauses can be redefined. For instance, by adjusting the left-right order of points on line BD, ``line B C D" is equivalent to ``line D C B"; altering the clockwise order of points on circle A, ``$\odot$A lieson E B D" and ``$\odot$A lieson E D B" are equivalent.
    
    \item \textit{Representation Transposition}: The geometric primitives, e.g., lines, angles and arcs, have multiple equivalent representations. For example, ``EA = AE", ``$\angle$STR = $\angle$RTS", ``$\widehat{\rm{EF}}$ = $\widehat{\rm{FE}}$". Randomly transposition of geometric primitive representations increases the diversity of problem representations.
    
    \item \textit{Clauses Shuffle}: The relative order of semantic clauses is randomly shuffled to generate new variable IDs, while modifying corresponding solution program. For example, when the semantic clauses are adjusted to ``CA = 3(N0), BD $\perp$ EA on C, EC = 2(N1)", the solution program is adjusted to ``Sum N0 N1 V0 Gougu N0 V1 V0 Multiple V1 C2 V2 Get V2".
    
    \item \textit{Diagram Flip}: The visual texts in diagram could be largely disregarded as they have been parsed into the fine-grained textual clauses. Therefore, a diagram that are flipped or rotated is considered equivalent to the original diagram, but this could enhance the multiplicity of global visual representation.
\end{itemize}

In conclusion, these five augmentation strategies are mutually independent yet can be synergistically integrated. The substantial number of samples generated by problem augmentation endow the neural solver with the primary geometric representation knowledge, thereby facilitating the high-level geometric reasoning.

\section{PGPSNet-v2 Solver}

\subsection{Preliminary and Pipeline}

\begin{figure}[t]
    \centering
    \includegraphics[width=0.49\textwidth]{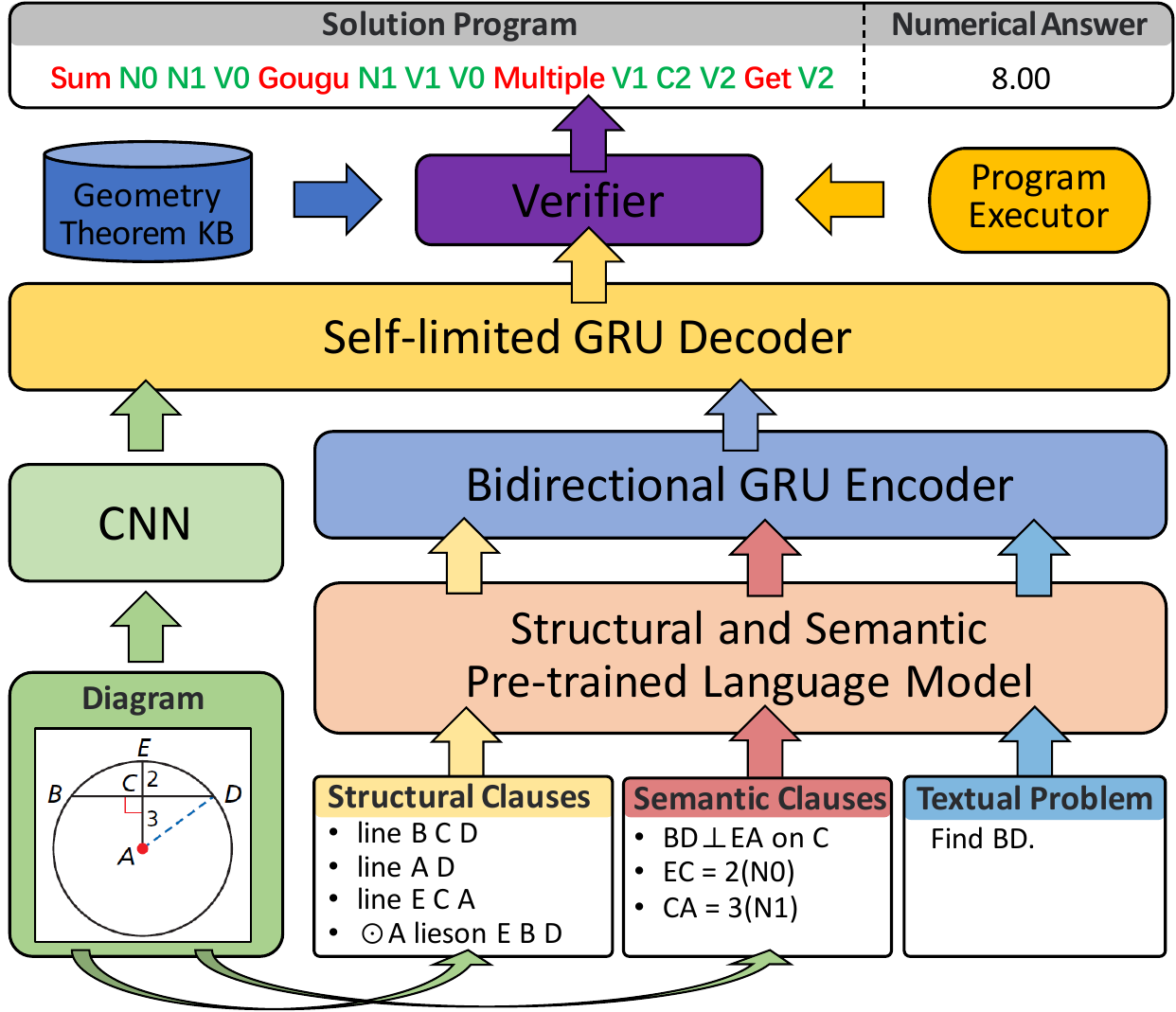} 
    \caption{Pipeline of PGPSNet-v2 solver.}
    \label{fig:PGPSNet}
\end{figure}

We begin by presenting the formal definition of GPS task. For a given geometry problem $P$, comprising a geometry diagram $D$ and a textual problem $T_{prob}$, the objective is to solve the problem and generate the solution $\mathcal{S}$, formulated as $P\!=\!\{D, T_{prob}\} \rightarrow \mathcal{S}$. In our work, GPS process is refined into three core steps: fusion, reasoning and verification, and it is re-expressed as:
\begin{equation}
    P \xrightarrow{\text{Fusion}} P' \xrightarrow{\text{Reasoning}}  \{\mathcal{S}\} \xrightarrow{\text{Verification}}  \mathcal{S'},
\end{equation}
where $P'$ is the new problem representation after modal fusion with $P$; Through neural reasoning for $P'$, it acquires a list of solution candidates $\{\mathcal{S}\}$; The solver gets the final solution $\mathcal{S'}$ via knowledge verification to $\{\mathcal{S}\}$.

Improved on above three core steps, we propose a neural-symbolic geometric solver, called PGPSNet-v2, which is extended from our previous neural solver PGPSNet~\cite{Zhang2023-ijcai} by adding a knowledge verifier, as displayed in Figure \ref{fig:PGPSNet}. The inputs of PGPSNet-v2 include not only the geometry diagram image $D$ and the textual problem $T_{prob}$, but also the structural clauses $T_{stru}$ and the semantic clauses $T_{sem}$ parsed from the geometry diagram. So, the problem text is expanded to be expressed as $T\!=\!\{T_{stru}, T_{sem}, T_{prob}\}\!=\!\{t_{j}\}_{j=1}^{N_T}$ after text concatenation, where $N_T$ is the text token number. In the modal fusion phase, the geometry diagram is encoded by a convolutional neural network (CNN) module and then flattened into a feature sequence $D'\!=\!\{h^{D}_i\}_{i=1}^{N_D}$, where $N_D$ is the diagram token number. In parallel, the problem text undergoes fused encoding through the structural-semantic pre-trained language model and the bidirectional GRU encoder in turn, and obtains a feature sequence $T'\!=\!\{h^{T}_j\}_{j=1}^{N_T}$. During the reasoning stage, two types of modal tokens, concatenated and unified as $P'\!=\!\{D', T'\}\!=\!\{h^{E}_i\}_{i=1}^{N_D\!+\!N_T}$, are fed into the self-limited GRU decoder to perform geometric reasoning, and the decoder generates corresponding solution candidates represented by multi-step program sequences $\mathcal{S}\!=\!\{S_i\}_{i=1}^{N_K}\!=\!\{s_j\}_{j=1}^{N_M}$, where $N_K$ and $N_M$ are the numbers of solving steps and program tokens, respectively. As to verification, combined with the geometry theorem knowledge base and the program executor, the verifier validates all the solution candidates and determines the final solution program $\mathcal{S'}$ and its numerical answer. The key modules of pre-training, decoder and verifier are detailed in the following.

\subsection{Structural-Semantic Pre-training}

\begin{figure*}[t]
    \includegraphics[width=0.98 \textwidth]{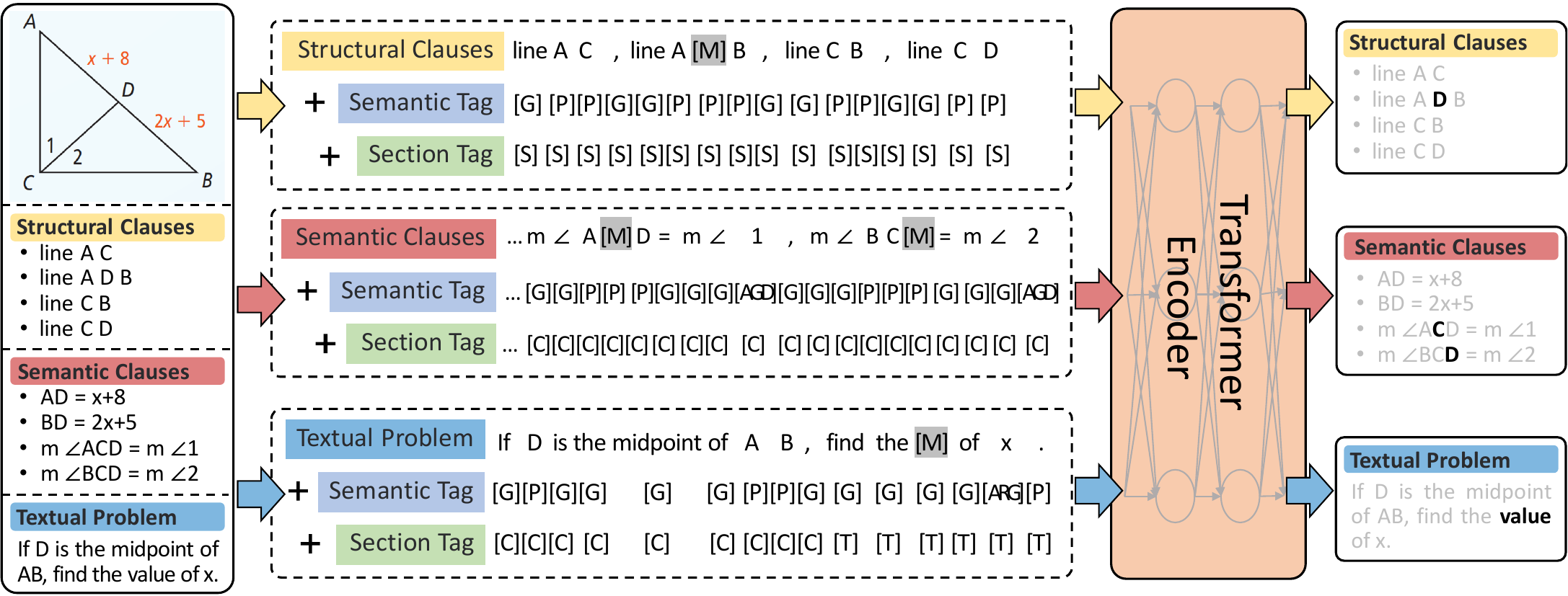} 
    \caption{Schematic flowchart of structural-semantic pre-training. [M] denotes the masked token. Semantic tags [G], [N], [ARG], [P], [AGD] represent tokens of general, variable, argument, point and angle ID. Section tags [S], [C], [T] refer to tokens of structure, condition and target.}
    \label{fig:Pre-trained_LM}
\end{figure*}

Structural-semantic pre-training plays a critical role in the modal fusion, by unifying textual clauses from diagram with the textual problem. While the textual clauses delineate fine-grained structural and semantic information extracted from diagram, they remain at low level, lacking overall coherence and contextual connections. Furthermore, verbose, fragmented and disorganized text continues to pose significant challenge to modal fusion and semantic structure comprehension. Inspired by pre-trained language models, as illustrated in Figure~\ref{fig:Pre-trained_LM}, we introduce the structural-semantic pre-training strategy based on the Masked Language Modeling (MLM) task~\cite{Devlin2019}, aiming to enhance structural and semantic understanding across modalities.

For pre-training, PGPSNet-v2 solver assigns semantic and section tags to each token in the problem text. The semantic tags refer to the semantic categories of tokens, encompassing general [G], variable [N], argument [ARG], point [P] and angle ID [ANGID]. The section tags indicate the portion to which a token belongs, with a given problem divided into three parts: structure [S], condition [C] and target [T]. The input textual token embedding $e(t_j)$ not only incorporates positional encoding but also integrates embeddings of semantic and section tags, formulated as:
\begin{equation}
    \begin{aligned}
    e(t_j) = & \text{TokenEmb}(t_j) + \text{PosEmb}(j) + \text{SemEmb}(t_j) \\
    &+ \text{SectEmb}(t_j), \;\; 1 \leq j \leq N_T.
    \end{aligned}
\end{equation}
The fine-grained semantic and section tags facilitate geometric modeling and also benefit in alleviating the imbalance among textual tokens. Subsequently, following the work of Cho et~al.~\cite{Cho2021}, we apply masked tokens [M] to obscure 30\% of textual tokens while keeping the semantic and section tags unchanged. The pre-training objective entails the restoration of masked textual tokens through a unified text generation approach~\cite{Devlin2019}.

Via the fused modal representation, this pre-training strategy proves highly applicable for fusing structural and semantic information. For instance, according to the clause ``D is the midpoint of AB" in the textual problem, it allows the inference that the masked token in the structural clause ``line A [M] B" is ``D", thereby promoting the model to acquire the geometry knowledge about point arrangement on lines. Similarly, considering the structural clauses ``line A C" and ``line C D", it can be deduced that the masked token in the semantic clause ``m $\angle$A[M]D = m $\angle$1" is ``C", by enhancing the model's understanding of geometric concept related to two lines intersecting at a point. Nevertheless, sometimes masked tokens cannot be inferred accurately, but the candidate tokens still encapsulate rich geometry knowledge. Taking the semantic clause ``m $\angle$BC[M] = m $\angle$2" as an example, based on the structural clauses ``line A C", ``line C B" and ``line C D", it is evident that the masked token is likely to be either ``D" or ``A". 

In summary, the structural-semantic pre-training contributes to fuse geometry problem representation via local relationship modeling. It endows PGPSNet-v2 solver with multi-modal geometric cognition capability, constituting the foundation for following geometric reasoning.

\subsection{Self-limited Decoder}

\begin{figure*}[t]
    \centering
    \includegraphics[width=0.78\textwidth]{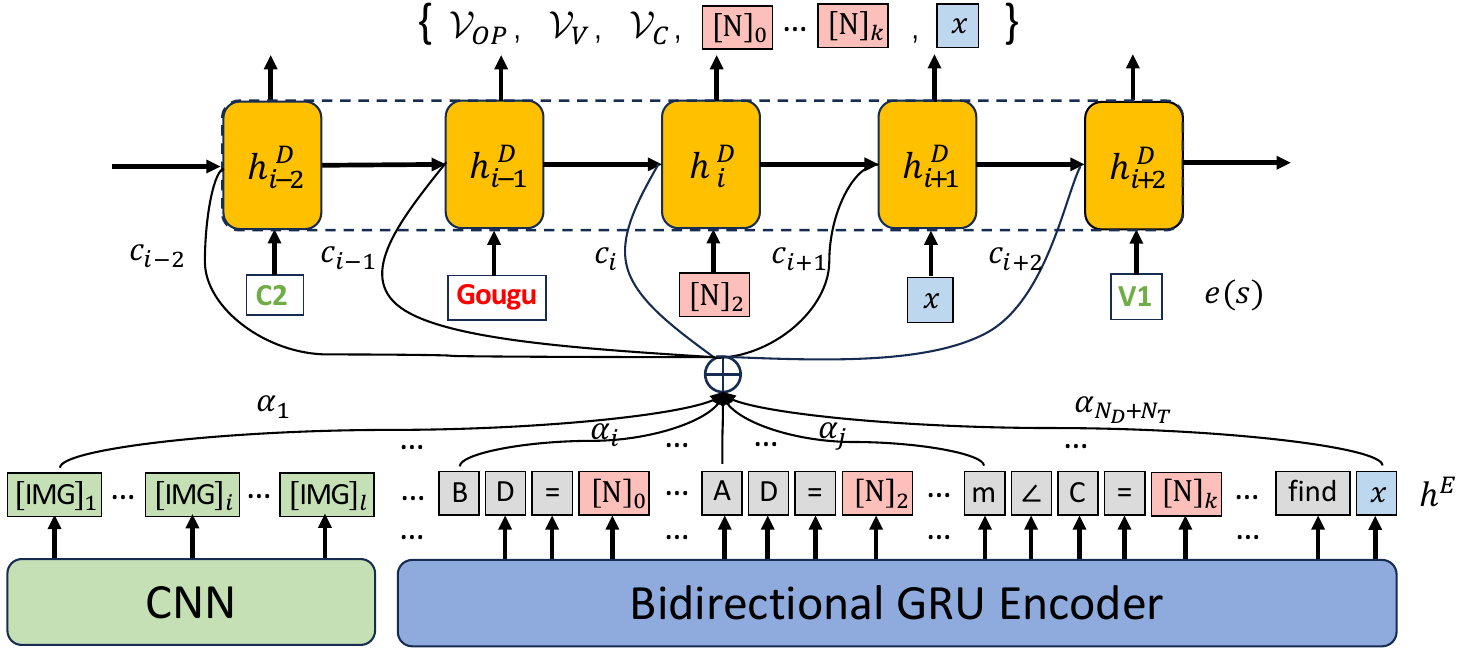}
    \caption{Schematic flowchart of self-limited decoder.}
    \label{fig:limited-decoder}
\end{figure*}

The self-limited decoder incorporates diagram and problem text to implement multi-modal geometric reasoning. In the reasoning process, the problem text enhanced by modal fusion provides rich structure and semantic information, and the diagram encoded by a CNN module further complements visual spatial content. Considering the complexity and flexibility of solving process of geometry problems, the solution program cannot convert into a binary or general expression tree, so the tree decoder widely used in math word problem (MWP) task~\cite{Xie2019, Tsai2021} is not applicable to GPS. As shown in Figure~\ref{fig:limited-decoder}, our self-limited GRU decoder is an attention-based decoder~\cite{Bahdanau2015} that generates the sequential solution program in an autoregressive manner, and boosts in two aspects:

(1) \textit{Reducing the complexity of input feature space}. In self-limited decoder, the input embeddings of problem variables $N$ and augments $ARG$ are copied from encoding context $\{h^{E}_i\}_{i=1}^{N_D\!+\!N_T}$ produced by the fusion module, based on the fact that problem variables and augments in solution program also exist in the problem text uniquely. This copying mechanism not only reduces the complexity of input feature space, but also enriches decoder inputs with contextual information. Specifically, the token embeddings of decoder inputs are defined as
\begin{equation}
    e(s) = \left\{
    \begin{aligned}
        &\text{TokenEmb}(s), \quad s \in \{\mathcal{V}_{O\!P}, \mathcal{V}_V, \mathcal{V}_C\}, \\
        &h^{E}_{loc(s, T)}, \qquad \qquad  s \in \{\mathcal{V}_N, \mathcal{V}_{A\!R\!G}\}, \\
    \end{aligned}
    \right.
\end{equation}
where $\mathcal{V}_{O\!P}$, $\mathcal{V}_V$, $\mathcal{V}_C$, $\mathcal{V}_{N}$ and $\mathcal{V}_{ARG}$ are the target vocabularies of operators, intermediate variables, constants, problem variables and augments, respectively; $loc(s,T)$ is the location of program token $s$ in the problem text $T$.

(2) \textit{Narrowing the search space of output token}. Self-limited decoder limits the output token candidates of problem variables $N$ and augments $ARG$ into that appear in the problem text $T$. Concretely, the probability of predicted token $s$ is
\begin{equation}
       \mathcal{P}(s) = \text{Softmax}(\text{Score}(h^D, c, e(s)))
\end{equation}
where $s\!\in\!\{\mathcal{V}_{O\!P}, \mathcal{V}_V, \mathcal{V}_C, \mathcal{V}_N \cap T, \mathcal{V}_{ARG} \cap T\}$, $h^D$ is the hidden vector for decoder; $c$ is the context vector generated from $h^E$ using the same attention mechanism as~\cite{Bahdanau2015}; The specific form of score function is $\text{Score}(h^D, c, e(s))\!=\!W_{\alpha}^T\text{tanh}\left(W_{\beta}^T\left[h^D\,\|\,c\,\|\,e(s)\right]\right)$, $W_{\beta}$ and $W_{\alpha}$ are learnable matrix parameters.

In our experiment, it was observed that the self-limited decoder achieved superior reasoning performance with faster training and inference speeds compared to sophisticated tree decoders, also yielding comparable results in MWP task~\cite{Xie2019}.

\subsection{Multi-level Theorem Verification}

Data-driven based neural solvers are susceptible to the size and content distribution of problem datasets, often generating solution process with high confidence but inconsistent with geometric principles. To alleviate this issue, together with the knowledge base and the program executor, we propose a geometry theorem knowledge verifier. Benefiting from the interpretable solution program design of PGPS9K dataset, the verifier validates the solution program generated by the self-limited decoder from three levels: form, calculability and semantic, as illustrated in Figure \ref{fig:verifier}:

\begin{figure*}[t]
    \centering
    \includegraphics[width=0.81 \textwidth]{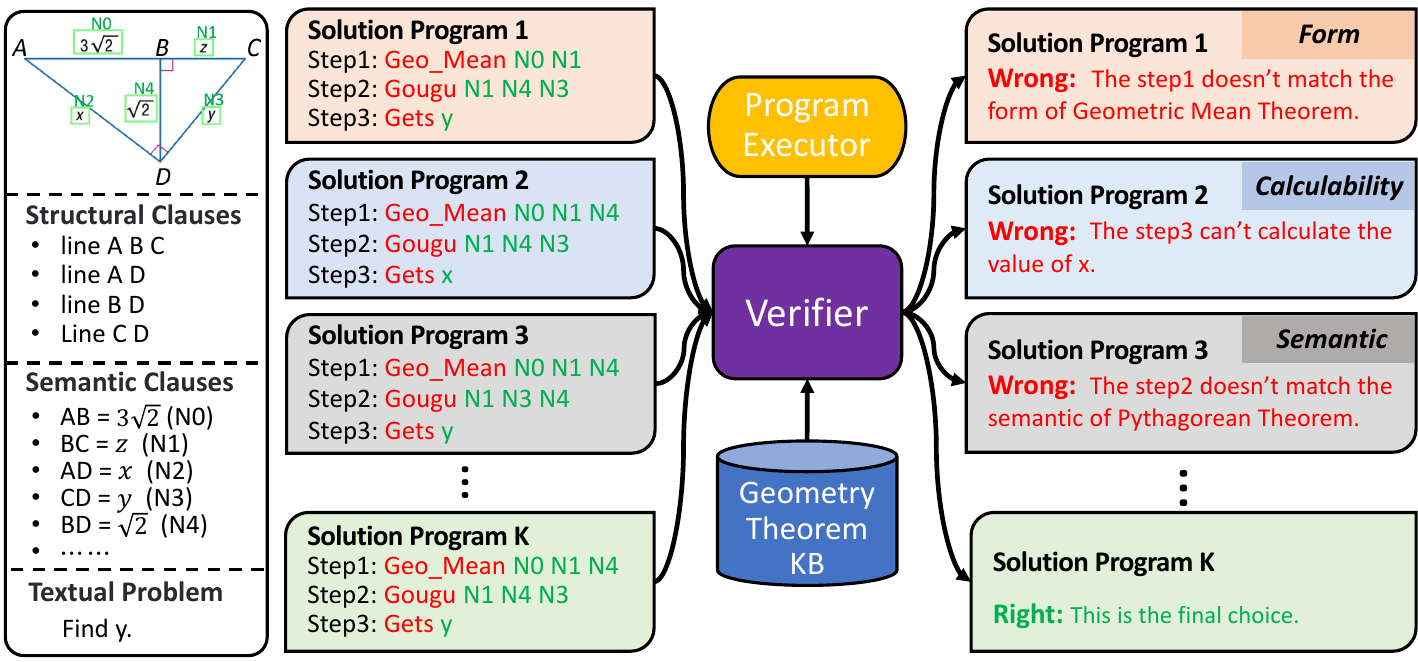}
    \caption{Multi-level knowledge verifier.}
    \label{fig:verifier}
\end{figure*}

\vspace{-0.2cm}

\begin{itemize}
    \item \textit{Form}: In the form level, it necessitates the solving step $S_i=(O\!P', P\!N')$ conform to the structure of knowledge tuple, which also means that $P\!N'$ aligns with the operand quantity of knowledge tuple. For instance, the solving step 1 of solution program 1, denoted as ``Geo\_Mean N0 N1", fails to adhere to the form of Geometric Mean Theorem that should express as ``Geo\_Mean(a, b, c)" instead.
    
    \item \textit{Calculability}: The calculability ensures that the solving steps $S_i$ are effectively computable according to the theorem formulas of knowledge tuples. For example, the solving step 3 of solution program 2, namely ``Get x", cannot compute the value of x.
    
    \item \textit{Semantic}: The semantic level requires that the operands $P\!N$ of solving step $S_i$ satisfy the semantic rules specified by the knowledge tuple. Taking the solving step 2 of solution program 3 as an example, ``GouGu N1 N3 N4" does not meet the semantic rules of Pythagorean Theorem that N1 and N3 are right legs and N4 is corresponding hypotenuse.
\end{itemize}

\begin{algorithm}[htbp]
    % \small
	\caption{Multi-level Theorem Verification.}
	\label{alg:algorithm1}
	\KwIn{Solution candidates $\{S\}$; Program executor $Q$; Theorem knowledge base $\mathcal{KB}$: the collection of knowledge tuples $K\!T = (O\!P, P\!N, T\!F, S\!R)$, where $O\!P, P\!N, T\!F, S\!R$ are operator, operands, theorem formula and semantic rules.}
	\KwOut{Selected solution $\mathcal{S}'$.}  
	\BlankLine

	\For{$solution~\mathcal{S}~\textbf{in}~\{\mathcal{S}\}$}
	    {
	        \For{$step~S_i=(O\!P', P\!N')~\textbf{in}~\mathcal{S}$}
	        {
	            \textbf{Base search}: Search $K\!T$ by $O\!P'$ in $\mathcal{KB}$;
	            
                \If {$K\!T$ isn't retrieved} {
                    \textbf{break};
                }
                
                \textbf{Tuple matching (Form)}: match $P\!N'$ with $P\!N$;
                
                \If {form mismatch} {
                    \textbf{break};
                }
                
                \textbf{Tuple matching (Calculability)}: calculate $S_i$ according to $T\!F$;
                
                \If {incalculable} {
                    \textbf{break};
                }
                
                \textbf{Tuple matching (Semantic)}: verify $S_i$ according to $S\!R$;
                
                \If {semantic inconsistency} {
                    \textbf{break};
                }
	        }
	        \If {all verifications have passed } {
	            $\mathcal{S}'=\mathcal{S}$;
	            
                \textbf{break};
            }
		}
	\textbf{return} $\mathcal{S}'$.
\end{algorithm}

\vspace{-0.5cm}

The hierarchical relationships among these three levels of verifications can be indicated as ``$Form \subset Calculability \subset Semantic$". In other words, if the form of solving step is incorrect, it is definitely computationally infeasible and semantically erroneous; if the solving step cannot be calculated, there also exists semantic issues. In addition, the verification complexities of the three levels are also increasing accordingly, in which form matching is the simplest and the semantic judgement is the most complex. The algorithmic workflow of multi-level theorem verification is demonstrated in Algorithm~\ref{alg:algorithm1}, primarily consisting of two steps: (1) \textit{Base Search}: Retrieving corresponding knowledge tuples from the theorem knowledge base based on the operator of solving step; (2) \textit{Tuple Matching}: Sequentially validating solving steps in three levels of form, calculability and semantic. During the verification process, the sub-loop terminates prematurely if any unsatisfactory verification arises. Among the solution programs that survive all the verification steps, the one with the highest confidence is selected as the final solution.

\section{Experiments}

\subsection{Experimental Settings}

\subsubsection{Implementation Details}

Our PGPSNet-v2 solver is implemented on the PyTorch framework using four 24GB NVIDIA GTX-RTX GPUs. The neural model adopts ResNet-18 \cite{Shafiq2022} as the CNN module, scales geometry diagrams into images of size $224*224$, and outputs flattened sequence of length $N_D=7*7$. We select three different scales of general transformer encoder~\cite{Vaswani2017} as the pre-trained language model architecture, signified as PGPSNet-S/M/L for solvers, which have 6/6/12 modules with each module containing 8/8/12 self-attention heads. The input embedding and hidden embedding in the models of three scales have dimensionality 256/512/768 and dimensionality 1,024/2,048/3,072, respectively. The GRU encoder is a two-layer bidirectional GRU~\cite{Cho2014} with the same dimensionality of input embedding and hidden state, aligned with the input embedding dimensionality of the pre-trained language model. The self-limited decoder is a two-layer GRU decoder~\cite{Cho2014} with both input embedding and hidden state in the same dimensionality as the GRU encoder. 

In training, we choose AdamW~\cite{Loshchilov2019} as the optimizer of PGPSNet-v2, with a regularization with weight coefficient of $1e^{-2}$ and a step-decline schedule with decaying rate of 0.5. During the pre-training stage, the learning rate of language model is initialized to $5e^{-4}$ and decays at 1K, 2K and 3K epochs, with a total of 4k epochs. During the training stage, all modules of PGPSNet-v2 are jointly trained. The initial learning rate of the language model is set as $1e^{-4}$ while the other modules are set as $1e^{-3}$, and all modules are uniformly decayed at 140, 280, 360, 440 and 500 epochs, in total 560 epochs. Additionally, the training batch size and dropout rate are set as 128 and 0.2, respectively. The problem augmentation is carried out synchronously in model training, and the augmentation probability is set as 0.7 and 0.5 for pre-training and fine training, respectively.

\subsubsection{Datasets and Evaluation}

We split the PGPS9K dataset into a training set and a test set in two distinct ways, signified as Geometry3K and PGPS9K in the following experiments: The first way takes the test samples of Geometry3K dataset~\cite{Lu2021} as the test set (589 samples), with the remaining non-overlapping samples constituting the training set (8,432 samples); The second way partitions the dataset based on problem types and allocates them in an 8:1 ratio, resulting in 8,021 samples for training and 1,000 samples for testing. For the convenience of experimental comparison, we leverage geometry diagram annotations to generate the formal language for the symbolic solvers and textual clauses for the neural solvers and our PGPSNet-v2. Considering the lack of suitable geometric corpus, coupled with the substantial disparity from natural language corpora, our language models are per-trained from scratch on the PGPS9K dataset with the solution programs masked. GPS datasets such as GeoQA~\cite{Chen2021} and GeoQA+~\cite{Cao2022} are not utilized or compared in our experiments, as they lack fine-grained diagram annotations and exhibit low-quality images, hard to be parsed by existing diagram parsers~\cite{Lu2021,Zhang2022}.

The performance of GPS is assessed in two aspects: numerical answer and solution program. For each aspect, three evaluation modes are employed: \textit{Completion, Choice and Top-3}. In the \textit{Completion} mode, the symbolic solver provides search results while the neural solver selects the best solution sequence. The \textit{Choice} mode is defined as choosing the correct option from four candidate answers, and one is randomly chosen if the answer is not among them. In the \textit{Top-3} mode, it is considered correct if the ground truth is among the top three candidate solutions, offering a relatively loose criterion compared to the Completion. Consistent with the experimental setup of~\cite{Chen2021,Chen2022}, our solvers set the beam size as 10 for all evaluation modes. It is noteworthy that the evaluation results of solution program often underestimate the actual performance due to the equivalence of operation orders and the diversity of solution strategies.

\subsection{Comparison with State-of-the-art Solvers}

We compare with state-of-the-art solvers including symbolic solvers and neural solvers to show the superior problem solving performance of PGPSNet-v2 as illustrated in Table~\ref{tab:solve_performance_new}.

For symbolic solvers, Inter-GPS~\cite{Lu2021} solved geometry problems by searching and matching with unified formal language. According to the input source of formal language, Inter-GPS presents three types of results, e.g., “Predict” means that all formal clauses are predicted by its parsers, “Digram GT” denotes that formal clauses of diagram are from the ground truth, and “All GT” indicates that formal clauses of diagram and textual problem are all ground truth. GeoDRL~\cite{Peng2023} improved the search strategy of Inter-GPS with logical graph deduction and deep reinforcement learning. Noting that the numbers of parameters of InterGPS and Geoformer are aligned with PGPSNet-v2-L but are much more than the other scales of PGPSNet-v2. Experimental results show that the PGPSNet-v2 outperforms symbolic solvers on two datasets and in almost evaluation metrics. Even compared with Inter-GPS (All GT) that uses annotated formal clauses designed carefully, PGPSNet-v2-L gains a 3.8\% improvement in Completion and a 5.3\% improvement in Choice on Geometry3K.

\begin{table*}[t]
\centering
\caption{Performance comparison of geometric problem solvers. * denotes results re-produced with the authors' code. \# denotes methods re-implemented by us. $T$ refers to the version without knowledge verification (i.e. PGPSNet), while the other versions are the variants of PGPSNet-v2.}
\begin{tabular}{lcccccc}
    \toprule
    \multirow{2}{*}{Model}  & \multicolumn{3}{c}{Geometry3K} & \multicolumn{3}{c}{PGPS9K} \\
    \cmidrule(lr){2-4} \cmidrule(lr){5-7}
                 & Completion & Choice & Top-3 & Completion & Choice & Top-3  \\
    \midrule
    Human Expert \cite{Lu2021} & - & 90.9 & - & - & - & - \\
    Baseline (Neural Solver) \cite{Lu2021} & - & 35.9 & -  & - & - & -  \\
    Inter-GPS (Predict)* \cite{Lu2021} & 44.6  & 56.9  & -  & -   & -  & -  \\
    Inter-GPS (Diagram GT)* \cite{Lu2021} & 64.2  & 71.7 & - & 59.8 & 68.0 & - \\
    Inter-GPS (All GT)* \cite{Lu2021} & 69.0 & 75.9 & - & -  & -  & - \\
    GeoDRL (Predict) \cite{Peng2023} & - & 68.4 & - & -  & -  & - \\
    NGS$^\#$ \cite{Chen2021} & 35.3 & 58.8 & 62.0 & 34.1 & 46.1 & 60.9 \\
    Geoformer$^\#$ \cite{Chen2022} & 36.8 & 59.3 & 62.5 & 35.6 & 47.3 & 62.3 \\
    SCA-GPS \cite{Ning2023} & - & 76.7 & -  & - & - & -   \\
    \midrule
    PGPSNet-v2-S (Predict)  & 65.2 & 76.4 & 77.6  & 60.3 & 69.2 & 78.1  \\
    PGPSNet-S (Clause GT)$^T$ & 65.8 & 78.6 & 80.1  & 59.7 & 69.5 & 80.1 \\
    PGPSNet-v2-S (Clause GT)  & 68.2 & 79.8 & 81.6  & 63.4 & 71.7 & 81.2    \\
    PGPSNet-v2-M (Clause GT)  & 70.7 & 80.0 & 82.2  & 66.9 & 74.3 & 81.5    \\
    PGPSNet-v2-L (Clause GT)  & \textbf{72.8} & \textbf{81.2} & \textbf{83.5}  & \textbf{69.4} & \textbf{75.8} & \textbf{82.9}    \\
    \bottomrule
\end{tabular}

\label{tab:solve_performance_new}
\end{table*}

As to neural solvers, NGS~\cite{Chen2021} and Geoformer~\cite{Chen2022} relied primarily on the textual problem to solve problems. Even through re-implementing them with the annotated textual clauses parsed from diagram and the same augmentation strategies, performance gaps between these two solvers and our PGPSNet-v2 are still significant, 37.5\% and 36.0\% lower than PGPSNet-v2-L in Completion on PGPS9K, respectively. SCA-GPS~\cite{Ning2023} shows similar performance as InterGPS (All GT) because its diagram understanding methods, character alignments and masked image modeling, are coarse-grained and ineffective. 

As to PGPSNet-v2, we also test the influence of textual clauses and parameter scales to GPS performance, e.g., “Predict” indicates that PGPSNet-v2 uses the textual clauses parsed by the PGDPNet parser~\cite{Zhang2022} instead of the ground truth, “T” denotes the version without knowledge verification (i.e., PGPSNet), and “S/M/L“ refers to different scales of language model. Compared PGPSNet-v2-S (Predict) with PGPSNet-v2-S (Clause GT), it turns out that the performance of two solvers is close for the strong parsing ability of PGDPNet. Knowledge verification brings an extra performance gain when comparing PGPSNet-S (Clause GT)$^T$ with PGPSNet-v2-S (Clause GT), which will be explained further in the ablation study. With the increase of model scale, the GPS performance is improved steadily, but saturates due to the limitation of data amount. In addition, the improvements of performance in Top-3 are less than that in
Completion because most of correct solutions are concentrated among top-rank candidates. There remains a certain performance gap between PGPSNet-v2 and human expert, however. This indicates automated GPS still has much room to improve in the future.

\subsection{Ablation Study}

To justify the effects of core steps in PGPSNet-v2 solver: modal fusion, reasoning process and knowledge verification, we conduct ablation studies by varying the combination of modules. 

\subsubsection{Effect of Fusion and Reason Modules}

For fusion and reason modules, we take self-limited decoder, problem augmentation, structural clauses, pre-trained language model and diagram attention as ablation objects to evaluate their effect on GPS. Table~\ref{tab:ablation_study} shows the ablation results of PGPSNet-v2-L on Geometry3K dataset. The comparison between row 1 and row 4 indicates that problem augmentation, by injecting geometric representation knowledge into augmented samples, benefits geometric logical reasoning. Through the comparison between row 2 and row 4, the results indicate that the self-limited decoder enhances the performance of GPS via simplifying the input feature space and restricting the search space, thereby reducing the difficulty of model learning. The language model with structural-semantic pre-training brings an amazing performance gain of 30.6\% accuracy improvement of numerical answer, as in the comparison between row 4 and row 6. Comparing row 3 and row 4, the results reveal that structural clauses have a relatively small impact on GPS performance in the absence of pre-training. However, with structural-semantic pre-training, structural clauses lead to a substantial improvement in GPS, as in the comparison between row 5 and row 6. This demonstrates that fundamental connection relationships
can enhance the model’s cognition of geometric structures with a befitting
modal fusion approach, thus aiding geometric logical reasoning. Visual diagram with cross-modal attention also provides rich visual layout information and brings a moderate improvement of GPS performance, as in comparing row 6 and row 7. 

\begin{table*}[t]
\centering
    \caption{Ablation study of fusion and reason modules on Geometry3K dataset.}
    \begin{tabular}{cccccccc}
        \toprule
        \multicolumn{5}{c}{Module} & \multicolumn{2}{c}{Accuracy}  \\
        \cmidrule(lr){1-5}  \cmidrule(lr){6-7}
        
        \makecell[c]{Self-limited \\ Decoder} & \makecell[c]{Problem \\ Augmentation} & \makecell[c]{Structural \\ Clauses} & \makecell[c]{Pre-trained \\ LM} & \makecell[c]{Diagram \\ Attention} & \makecell[c]{Numerical \\ Answer} & \makecell[c]{Solution \\ Program}  \\
        \midrule
        \ding{52} & \ding{56} & \ding{52} & \ding{56} & \ding{56} & 34.7 & 29.7   \\
        \ding{56} & \ding{52} & \ding{52} & \ding{56} & \ding{56} & 30.0 & 27.1   \\
        \ding{52} & \ding{52} & \ding{56} & \ding{56} & \ding{56} & 38.5 & 35.7   \\
        \ding{52} & \ding{52} & \ding{52} & \ding{56} & \ding{56} & 40.6 & 38.6  \\
        \ding{52} & \ding{52} & \ding{56} & \ding{52} & \ding{56} & 58.7 & 54.9  \\
        \ding{52} & \ding{52} & \ding{52} & \ding{52} & \ding{56} & 71.2 & 68.7  \\
        \ding{52} & \ding{52} & \ding{52} & \ding{52} & \ding{52} & \textbf{72.8} & \textbf{70.8} \\
        \bottomrule
    \end{tabular}

\label{tab:ablation_study}
\end{table*}

Additionally, in all the aforementioned experiments, the performance of solution program observes a consistent trend with numerical answer. Nevertheless, the performance of solution program is 2\%-5\% lower than that of numerical answer overall. This is due to the influence of the diversity of solution approaches for GPS.

\subsubsection{Impact of Verifier}

To justify the effect of knowledge verification at different levels, we conduct ablation experiments for the theorem knowledge verifier on PGPS9K dataset, with the evaluation mode of Completion in default. As shown in Table~\ref{tab:ablation_verify}, on the PGPSNet-v2-S solver, the introduction of form verification yields a 1.1\% improvement compared to the model without verification (i.e., PGPSNet-S), further incorporating calculability validation results in a 1.6\% enhancement relative to that with only form verification, and the inclusion of semantic verification contributes an additional 1.0\% improvement. Consequently, in comparison to the model without verifier, the ultimate improvement on PGPSNet-v2-S is 3.7\%. Furthermore, experimental results indicate that the theorem knowledge verifier all shows performance improvements across solvers of varying model scales. However, as the model scale increases, the extent of improvement diminishes. On PGPSNet-v2-M and PGPSNet-v2-L, the improvement rates are 3.0\% and 2.3\%, respectively. This is because the large-scale neural model without knowledge verification already performs fairly well. Nevertheless, the verifier still benefits the performance substantially.

\begin{table}[t]		
\centering	
\caption{Ablation studies of verifier on PGPS9K dataset. In the case of ``None'' verification, the PGPSNet-v2 model is equivalent to the neural model PGPSNet.}
\begin{tabular}{lccc}				
    \toprule				
    Verification Level & PGPSNet-v2-S & PGPSNet-v2-M & PGPSNet-v2-L \\			
    \midrule				
    None  & 59.7 & 63.9 & 67.1 \\				
    Form & 60.8 & 64.6 & 67.5 \\				
    Calculability & 62.4 & 66.0 & 68.7 \\				
    Semantic & \textbf{63.4} & \textbf{66.9} & \textbf{69.4} \\				
    \bottomrule				
\end{tabular}

\label{tab:ablation_verify}
\end{table}	

\vspace{-0.4cm}

\subsection{Case Analysis}

Detailed case analyses on PGPS9K dataset are conducted to discuss the capabilities and limitations of our PGPSNet-v2 solver. These case studies include the analysis of overall problem solving performance of PGPSNet-v2, and the ability test of theorem knowledge verifier.

\subsubsection{Problem Solving Cases}

\begin{figure*}[t]
    \centering
    \includegraphics[width=0.95\textwidth]{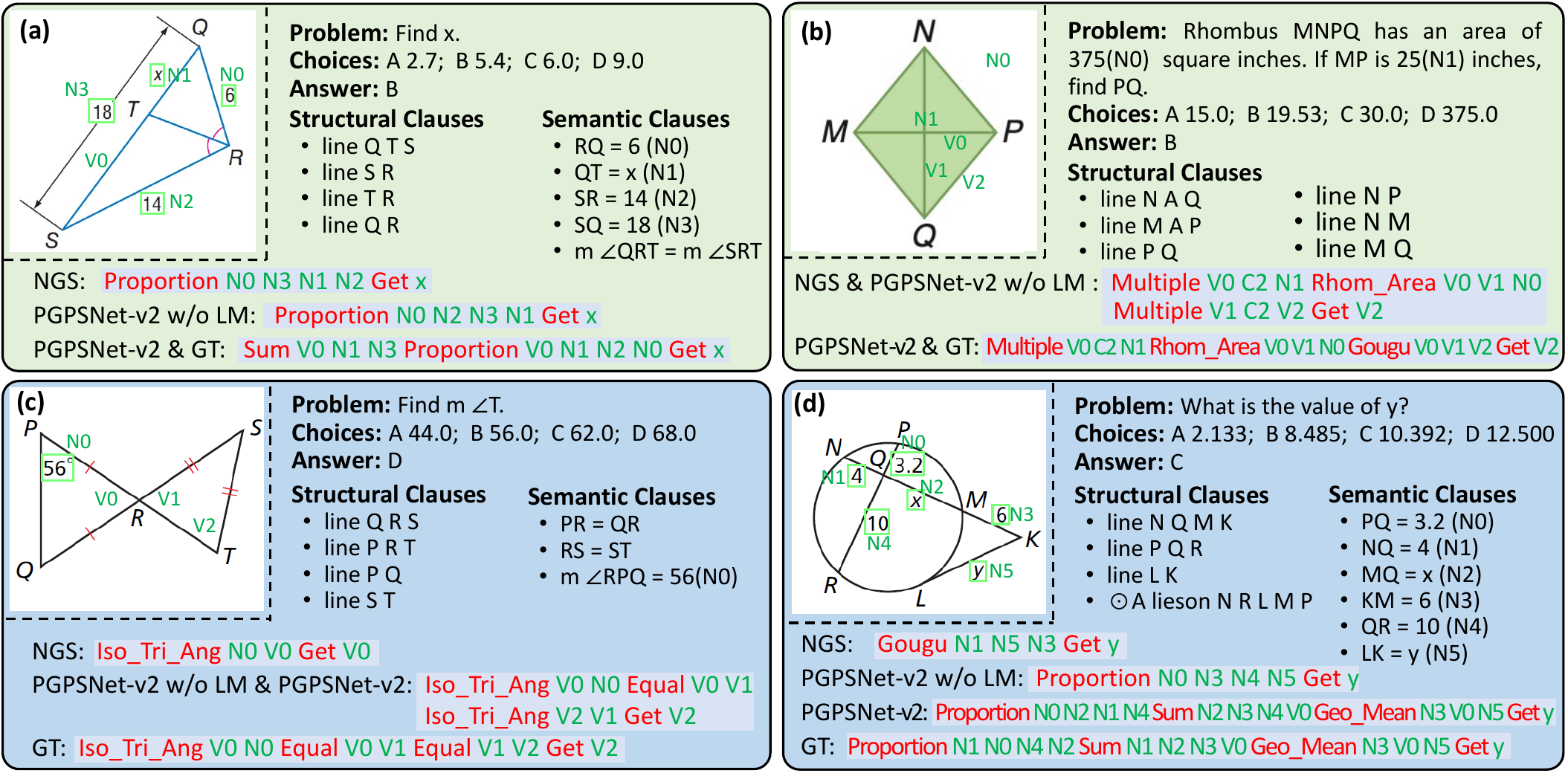}
    \caption{Case studies of problem solving. (a) and (b) are the problems PGPSNet-v2 answered correctly, (c) and (d) are the problems PGPSNet-v2 answered incorrectly.}
    \label{fig:case_study}
\end{figure*}

Figure \ref{fig:case_study} displays four plane geometry problems (a)-(d) and their solution programs of NGS \cite{Chen2021}, PGPSNet-v2 w/o LM, PGPSNet-v2 and ground truth, where (a) and (b) are the problems PGPSNet-v2 answered correctly while (c) and (d) are the problems PGPSNet-v2 answered incorrectly. Case (a) examines the application of Angle Bisector Theorem. Solvers NGS and PGPSNet-v2 w/o LM fail to address the proportional relationships between the side lengths of a triangle divided by its angle bisector, while PGPSNet-v2 generates the correct solution procedure; Case (b) investigates the knowledge point related to the calculation of rhombus area. In the penultimate step, both the NGS and PGPSNet-v2 w/o LM inaccurately provide the length of rhombus diagonal, whereas PGPSNet-v2 utilizes the Pythagorean Theorem correctly to compute the length of rhombus side; Case (c) necessitates the consideration of relationships among the interior angles of isosceles triangle, but all solvers struggle to effectively discern the relationships between base angle, apex angle and vertically opposite angle; Case (d) is a hard geometry problem that requires the application of two types of Chord-Length Theorems and involves multiple steps of theorem manipulation. In the context of problem (d), all solution programs from various solvers are proved to be incorrect, but the solution program generated by PGPSNet-v2 is demonstrated the closest solution to the ground truth. In summary, comprehensive experimental results suggest that PGPSNet-v2 outperforms than other solvers, showcasing significant potential for inferential cognition, though it still lack the capability for intricate geometric reasoning at present.

\subsubsection{Theorem Verification Cases}

\begin{figure*}[t]
    \centering
    \includegraphics[width=0.68 \textwidth]{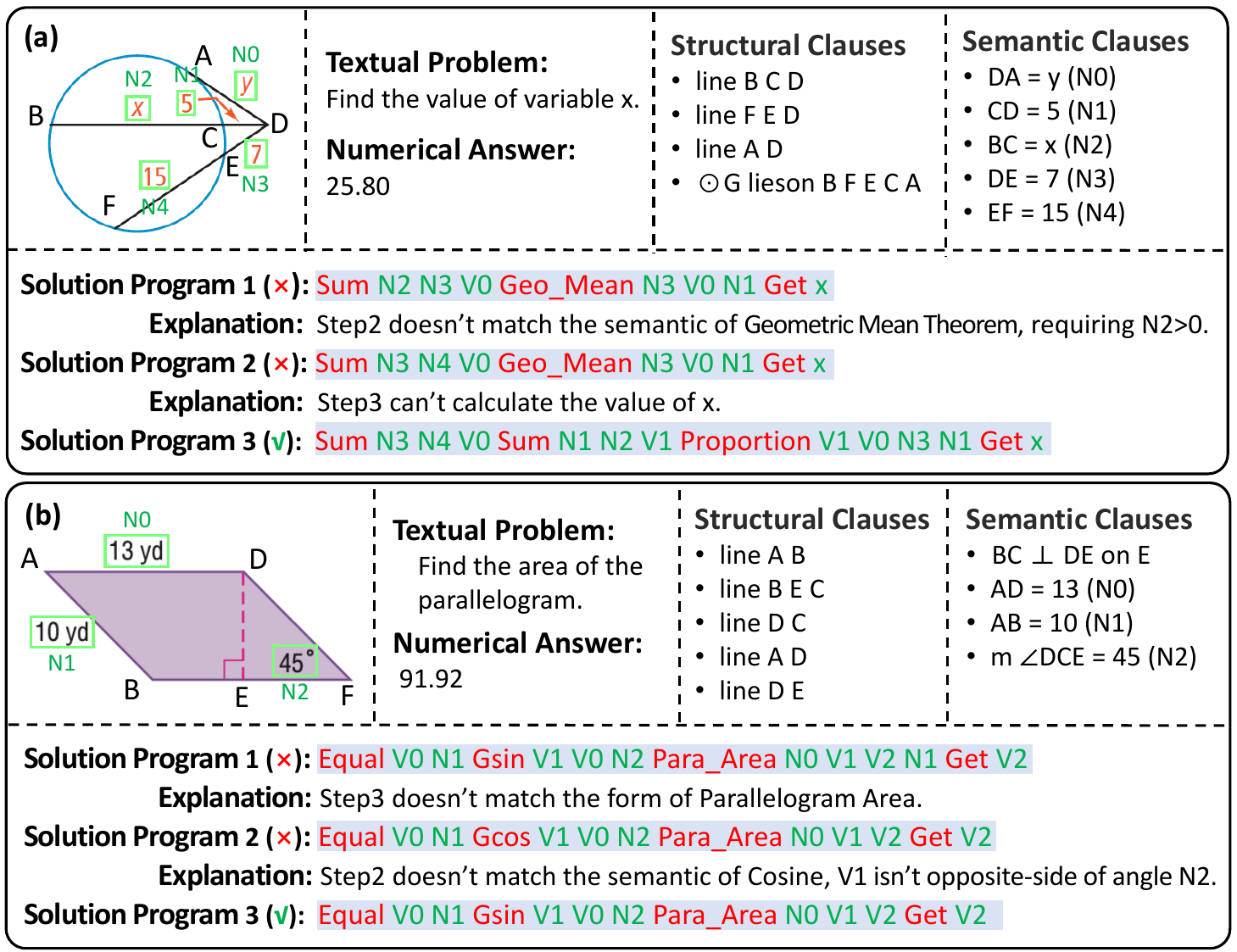} 
    \caption{Case studies of theorem knowledge verification.}
    \label{fig:verify_cases}
\end{figure*}

We demonstrate the pivotal role played by the theorem knowledge verifier of PGPSNet-v2 through two GPS cases, as shown in Figure \ref{fig:verify_cases}. Case (a) is a problem that examines the secant length relationship of circle, and the whole solving process consists of two main steps, first compute x and then calculate y, according to the Secant-Segment Theorem. The step 2 of solution program 1 gets the value of N2 less than zero, which does not meet the semantic rules of Geometric Mean Theorem, mainly because the calculation of line segment length in step 1 is wrong. When calculating the step 3 of solution program 2, due to lack of a clear problem solving goal for solver, the program executor acquires no value of x, violating the computability; Case (b) involves calculating the area of parallelogram and needs to determine the unknown height DE in terms of the hypotenuse AB and the diagonal angle $\angle$DFE. The step 3 of solution program 1 dissatisfies the form of parallelogram area formula ``Para\_Area(a, b, c)". Besides, the semantic rules of Cosine Law are not satisfied in step 2 of solution program 2, which requires V1 to be the diagonal side of N2. In short, the objective of theorem knowledge verifier is to screen solutions in the post-processing phase, aiming to compensate for the defects in ability of PGPSNet-v2 in the generation phase.

\subsection{Limitation Discussion}

Despite the significant progress achieved by our PGPSNet-v2 in GPS, it remains in nascent stage and is subject to certain limitations. Similar to conventional neural networks, the PGPSNet-v2 solver, consisting a neural model as main part, is data-hungry and needs extensive samples with robust annotations to facilitate effective model learning. However, in reality, a substantial portion of available geometry problems are unlabeled and low-quality. The textual clauses parsed from the geometry diagram cannot convey the content of diagram comprehensively. For instance, absolute spatial position and visual color information are lost and these features may play crucial roles in problem solving. In the modal fusion stage, PGPSNet-v2 extracts visual features with a CNN module in a coarse-grained way, inevitably causing information loss. In the process of reasoning, our solution program exhibits symbolic interpretability and controllability, but is not so flexible as natural language, rendering it challenging to manage with more advanced operations such as adding auxiliary points or lines. In terms of knowledge verification, PGPSNet-v2 currently adopts a strategy of post-processing the candidate solutions generated by the neural model, still depending on a substantial set of manually predefined knowledge rules. Moreover, the problem augmentation of PGPS9K does not bring the diversity of logical semantics, but only the expansion of geometric representations.

\section{Conclusion and Future Work}

In this paper, we present PGPSNet-v2, a neural-symbolic geometric solver that addresses GPS in three key steps: modal fusion, reasoning process, and knowledge verification. Textual clauses parsed from diagram are fused with textual problem via structural-semantic pre-training, and solution programs are generated though a self-limited decoder. A multi-level theorem verifier identifies and eliminates solutions that conflict with geometric principles, thereby reducing erroneous results and enhancing the explainability and reliability of solution. We also provide a comprehensive geometry problem dataset PGPS9K, featured with fine-grained annotations of textual clauses and solution programs, and a geometry knowledge base represented as knowledge tuples, which facilitates the research and evaluation of GPS. Extensive experiments on datasets Geometry3K and PGPS9K demonstrate the superiority of the proposed PGPSNet-v2, outperforming existing symbolic and neural solvers.

Future works can be conducted to further improve the fusion of modal information, considering better fusion model and cross-modal alignment mechanism, to incorporate theorem knowledge directly in the generation of solution process for better integrating neural and symbolic reasoning. There are also potential room of improvement via expanding the types of geometry problems, the annotated dataset, leveraging un-annotated data and large language models.

\bibliographystyle{IEEEtran}
\bibliography{ref}

% \newpage
\appendix 

\section{More Details about PGPS9K Dataset} 

\subsection{Dataset Distribution} \label{Problem Types}

\begin{figure*}
    \centering
    \includegraphics[width=0.9\textwidth]{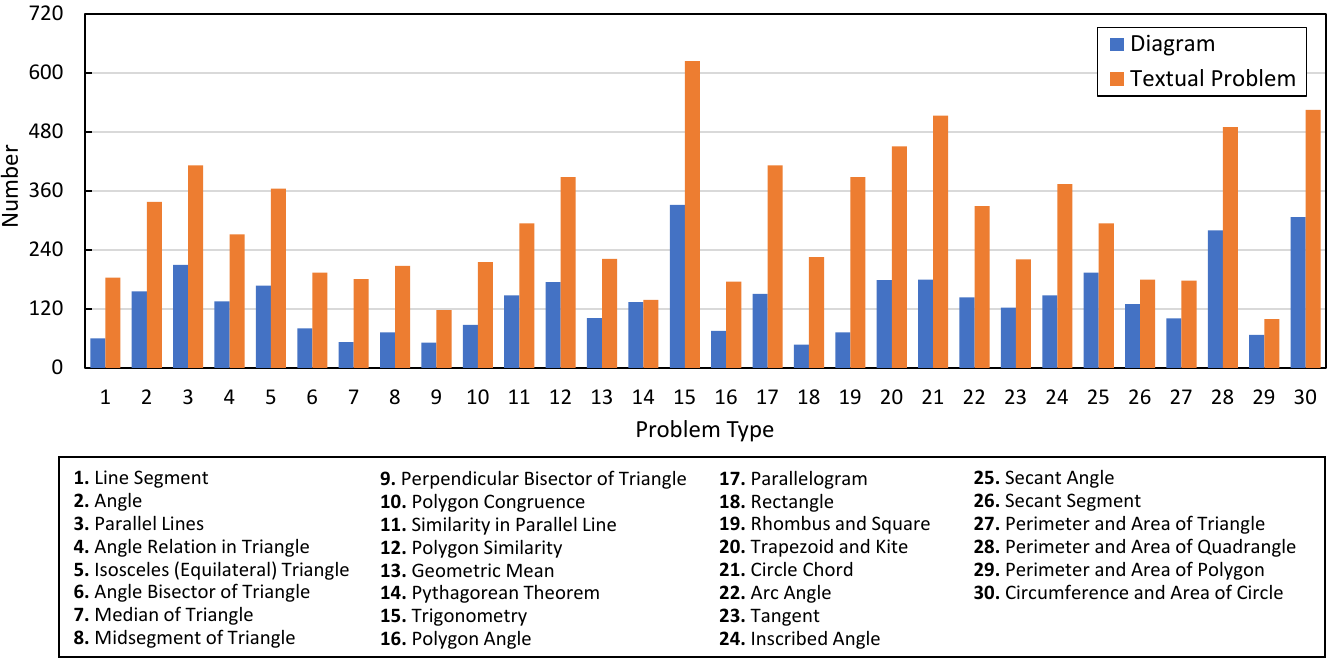} 
    \label{fig:problem_type}
    \caption{Problem type distribution of PGPS9K dataset.}
\end{figure*}

Our PGPS9K dataset is divided into 30 problem types elaborately according to geometry knowledge points by education experts, covering almost all problem types of plane geometry problem across grade 6-12 as shown in Figure~\ref{fig:problem_type}. Fine-grained partition helps geometry problem collection, trying to keep relative balance of problem types. For example, the poor performance of several problem types may result from insufficient problem samples, so we could collect them on purpose. Moreover, fine-grained partition is beneficial to analyse the model performance of different problem types in depth. For instance, there widely exists intersection and inclusion of geometric patterns among different problem types, and exploring the model capacity of inductive and deductive on GPS is meaningful and interesting.

\subsection{Textual Clauses} \label{Textual Clauses}

The textual clauses are the linguistic descriptions of primitive relations of geometry diagram. They include structural clauses and semantic clauses, where the structural clauses present the connection relations of geometric primitives (point, line and circle) and the semantic clauses depict the relations between geometric primitives and non-geometric primitives (text and symbol). Table \ref{tab:templates_clause} displays the complete templates of textual clauses, consisting 3 types of structural clauses and 6 types of semantic clauses. These textual clauses are elementary and necessary, which could be translated from relation tuples of diagram annotation directly without advanced geometric rules. The clause design is open, but neural solvers do not pursue high-level logical clauses though they may contribute to the problem solving.

\begin{table*}
    \caption{Templates of textual clauses. \&, *, \$ and \% denote point, line, variable and angle ID, respectively.}
    \centering
    \begin{tabular}{l|l|l|l} 
        \toprule
        Type	& Template	& Example & Explanation \\ 
        \midrule
        \multirow{4}{*}{\makecell[c]{Structural Clauses \\ (3 types)}} & line \& \& \& $\cdots$ & line A B C & \multirow{2}{*}{Points lie on line instances}  \\ 
        \cmidrule{2-3}
        & line * lieson \& \& \& $\cdots$ & line k lieson A B C & \\ 
        \cmidrule{2-4}
        & $\odot$\& lieson \& \& \& $\cdots$ & $\odot$O lieson E F G & Points lie on circle instances \\
        \midrule
        \multirow{9}{*}{\makecell[c]{Semantic Clauses \\ (6 types)}}   &  \&\& = \&\& = $\cdots$ = \$ & AB = CD = $3x+y$ & Length of line segments \\ 
        \cmidrule{2-4}
        & l $\widehat{\&\&}$  = l $\widehat{\&\&\&}$ = $\cdots$ = \$ & l $\widehat{\rm{EF}}$ = $5\pi$ & Length of arcs \\ 
        \cmidrule{2-4}
        & m $\angle$\&\&\& = m $\angle$\& = m $\angle$\% = $\cdots$ = \$ & m $\angle$A = m $\angle 1$ = $30$ & Degree of angles \\ 
        \cmidrule{2-4}
        & m $\widehat{\&\&}$  = m $\widehat{\&\&\&}$ = $\cdots$ = \$  & m $\widehat{\rm{EFG}}$ = $270$ & Degree of arcs \\ 
        \cmidrule{2-4}
        & \&\&(line *) $\parallel$ \&\&(line *) $\parallel$ $\cdots$ ~  & line k $\parallel$ line m $\parallel$ EF & Parallel relation among lines \\ 
        \cmidrule{2-4}
        & \&\&(line *) $\perp$ \&\&(line *) on \& & EF $\perp$ GH on C & Perpendicular relation among lines  \\
        \bottomrule
    \end{tabular}
    \label{tab:templates_clause}
\end{table*}

\begin{table*}
\centering
    \caption{Element set of solution program.}
    \begin{tabular}{ll|l} 
    \toprule
    \multicolumn{2}{l|}{\textbf{Type}} & \textbf{Element Set} \\ 
    \midrule
    \multicolumn{2}{l|}{Operator (34)} & \makecell[l]{Get, Equal, Sum, Multiple, Ratio,
      Median, Gougu, Gsin, Gcos, Gtan, \\ 
      Sin\_Law, Cos\_Law, Iso\_Tri\_Ang, Proportion, Geo\_Mean, Chord2\_Ang, \\
      TanSec\_Ang, Tria\_BH\_Area, Tria\_SAS\_Area, PRK\_Perim, Para\_Area, \\
      Rect\_Area, Rhom\_Area, Kite\_Area,  Trap\_Area, Circle\_R\_Circum, \\
      Circle\_D\_Circum, Circle\_R\_Area, Circle\_D\_Area, ArcSeg\_Area, \\
      Ngon\_Angsum, RNgon\_B\_Area, RNgon\_L\_Area, RNgon\_H\_Area}  \\ 
    \midrule
    \multirow{5}{*}{Operand (55)} & Problem Variable (11) & N0, N1, N2, $\cdots$, N10\\
              \cmidrule(lr){2-2}  \cmidrule{3-3}          
              & Intermediate Variable (7) & V0, V1, V2, $\cdots$, V6 \\
              \cmidrule(lr){2-2}  \cmidrule{3-3}  
              & Augment (26) & a, b, c, $\cdots$, x, y, z \\
              \cmidrule(lr){2-2}  \cmidrule{3-3}  
              & Constant (11) & C0.5, C2, C3, C4, C5, C6, C8, C60, C90, C180, C360    \\
    \bottomrule
    \end{tabular}
\label{tab:program_set}
\end{table*}

\subsection{Solution Program} \label{Solution Program}

Our annotation of solution program possesses better flexibility and scalability with extensive theorem operators and variable operands. The detailed contents of program set are listed in Table \ref{tab:program_set}. It should indicate that the solution program for GPS still confronts similar issues as general math word problem \cite{liang2023generalizing}: (1) \textit{Uncertainty caused by exchangeable operands}: The operands in some theorem formulas are commutative, e.g., in Pythagorean Theorem with formula $a^2+b^2=c^2$, the two right legs are exchangeable. In our annotation, we normalize solution program via specifying two-level priority of commutative operands. The first is the type level that ``Augment $>$  Intermediate Variable $>$ Problem Variable $>$ Constant" and the second is the index level with positive order. (2) \textit{Uncertainty of equivalent step orders}: Solving steps are in no particular order sometimes. We manually keep the same pre-defined step order for the same problem type. (3) \textit{Multiple solution methods}: A part of geometry problems could be solved by multiple solution methods. We choose the solution method with the most concise solution program. 

\end{document}